\documentclass[journal]{IEEEtran}

\usepackage{amsmath,amsfonts,bm}

\def\eqref#1{equation~\ref{#1}}

\def\1{\bm{1}}

\def\vphi{{\bm{\phi}}}

\def\vx{{\bm{x}}}

\DeclareMathAlphabet{\mathsfit}{\encodingdefault}{\sfdefault}{m}{sl}
\SetMathAlphabet{\mathsfit}{bold}{\encodingdefault}{\sfdefault}{bx}{n}

\def\gX{{\mathcal{X}}}
\def\gY{{\mathcal{Y}}}

\newcommand{\E}{\mathbb{E}}

\usepackage{graphicx}
\usepackage[list=true]{subcaption}

\usepackage[ruled]{algorithm2e}
\usepackage{algorithmic}

\newcommand{\dataset}{\mathcal{D}}
\newcommand{\datasetl}{\mathcal{D}_1}
\newcommand{\datasetu}{\mathcal{D}_0}
\newcommand{\lab}{\ell}

\newcommand{\gXinfo}{\gX_\star}
\newcommand{\modelinit}{\widetilde{q}}

\usepackage{amsthm}
\theoremstyle{definition}
\newtheorem{definition}{Definition}

\begin{document}
	
\title{Reliable Semi-Supervised Learning \\when Labels are Missing at Random}


\author{Xiuming~Liu,
        Dave~Zachariah,
        Johan~W{\aa}gberg,
        Thomas~B.~Sch{\"o}n
\thanks{Manuscript submitted 24 October, 2019. This research was financially supported by the following funding agencies and projects: the Swedish Research Council via the project \emph{Towards self-adaptive and resilient networked sensing systems} (project number: 2017-04543); the Swedish Research Council via the project \emph{Counterfactual Prediction Methods for Heterogeneous Populations} (project number: 2018-05040); the Swedish Foundation for Strategic Research (SSF) via the project \emph{ASSEMBLE} (contract number: RIT15-0012); and the Swedish Research Council via the project \emph{NewLEADS - New Directions in Learning Dynamical Systems} (contract number: 621-2016-06079). We thank the Swedish Research Council and the Swedish Foundation for Strategic Research for their support. }
\thanks{X.~Liu, D.~Zachariah, J.~W{\aa}gberg, and T.~B.~Sch{\"o}n are with Department of Information Technology, Uppsala University, Uppsala, Sweden.}}

\maketitle

\begin{abstract}
Semi-supervised learning methods are motivated by the availability of large datasets with unlabeled features in addition to labeled data. Unlabeled data is, however, not guaranteed to improve classification performance and has in fact been reported to impair the performance in certain cases. A fundamental source of error arises from restrictive assumptions about the unlabeled features, which result in unreliable classifiers that underestimate their prediction error probabilities. In this paper, we develop a semi-supervised learning approach that relaxes such assumptions and is capable of providing classifiers that reliably quantify the label uncertainty. The approach is applicable using any generative model with a supervised learning algorithm. We illustrate the approach using both handwritten digit and cloth classification data where the labels are missing at random. 
\end{abstract}

\IEEEpeerreviewmaketitle

\section{Introduction}
The goal of a classifier is to predict the class label $y \in \mathcal{Y}$ of an object with features $\vx \in \mathcal{X}$. Supervised learning of classifiers requires recording data pairs $(\vx, y)$, but obtaining labels $y$ for every observed feature $\vx$ is a costly and/or time-consuming process. This limitation prohibits learning accurate classifiers in many scenarios. By contrast, obtaining unlabeled data $\vx$ alone is often considerably simpler than labeled data $(\vx, y)$. For instance, obtaining large samples of speech recordings or x-ray scans is substantially easier than providing an accurate label to each sample \cite{rajpurkar2017chexnet}. This motivates the development of semi-supervised methods that leverage large amounts of unlabeled data in addition to a more limited labeled dataset, denoted
$$\dataset_0 = \{ \vx_i \} \quad \text{and} \quad \dataset_1 = \{ (\vx_i, y_i) \},$$
respectively. That is, methods applicable to scenarios in which $|\dataset_0| \gg |\dataset_1|$. A fundamental question that arises then is \emph{when will semi-supervised learning work?} \cite{chapelle2006semi}. More precisely, under what conditions can the unlabeled data improve the performance of a learning algorithm?   

\subsection{Assumptions on the missing data}
Semi-supervised learning methods face a fundamental statistical question:  
\begin{quote}
        Under what assumptions can $\datasetu$ improve upon the prediction performance as compared using only $\datasetl$? 
\end{quote}
Without explicitly addressing this question the result might involve incorrect utilization of the unlabeled data, which consequently degrades rather than enhances prediction performance. For example, \cite{oliver2018realistic} recently concluded that many methods \textquotedblleft suffered when the unlabeled data came from different classes than the labeled data.\textquotedblright{} See also \cite{cozman2003semi} and \cite{krijthe2016pessimistic}. Missing data is a well-studied area in statistics \cite{little2014statistical}. To provide a general description of the statistical limitations in semi-supervised learning, we consider each feature/label pair to be drawn from an underlying data-generating distribution, 
$$(\vx, y) \sim p(\: \vx, y \: | \: \ell \:) \; = \;  p(y |\vx, \lab) \: p( \vx | \lab),$$
where $\lab \in \{ 0,1\}$ is an indicator denoting whether the class label $y$ is missing ($l = 0$) or observed ($l = 1$). Note that $p(y |\vx, \lab)$ quantifies the label uncertainty and plays a critical role in the classification task. There are three main scenarios in which unlabeled data $\dataset_0$ is obtained:

\begin{figure*}
	\centering
	\subcaptionbox{Labeled and unlabeled data as colored and gray, respectively, with classes 0 (circles) and 1 (crosses).  \label{fig:example_a}}[.3\linewidth]
	{\includegraphics[width=0.3\textwidth]{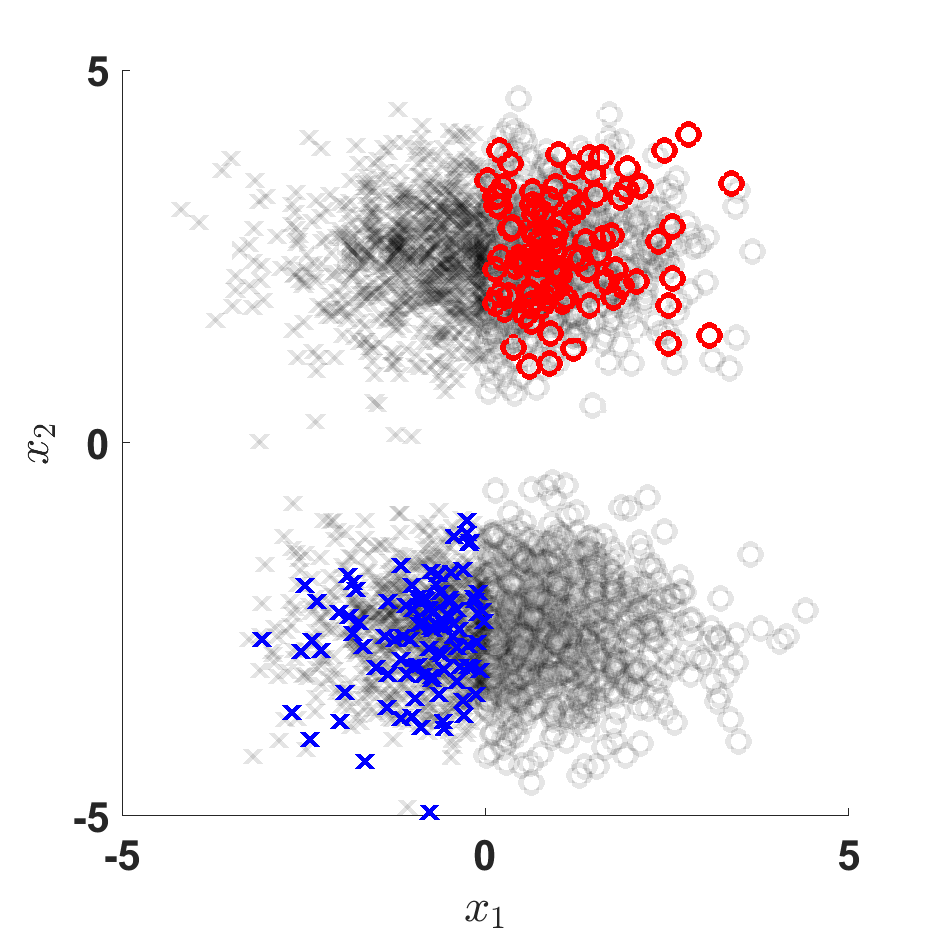}}%
	\hfill
	\subcaptionbox{Model of $p(y = 0\ |\ \vx)$ using MCAR-based semi-supervised learning methods.\label{fig:example_b}}[.3\linewidth]
	{\includegraphics[width=0.3\textwidth]{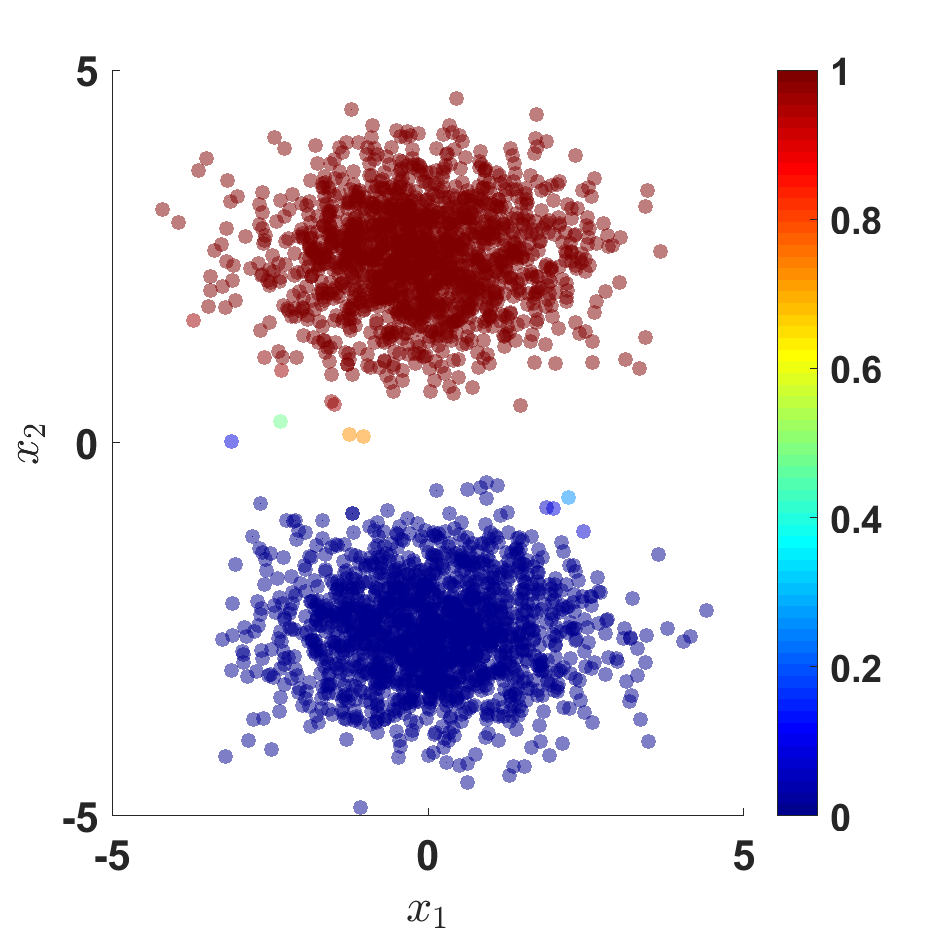}}%
	\hfill
	\subcaptionbox{Model of $p(y = 0\ |\ \vx)$ using reliable semi-supervised learning method described in Section~\ref{sec:method}. \label{fig:example_c}}[.32\linewidth]
	{\includegraphics[width=0.3\textwidth]{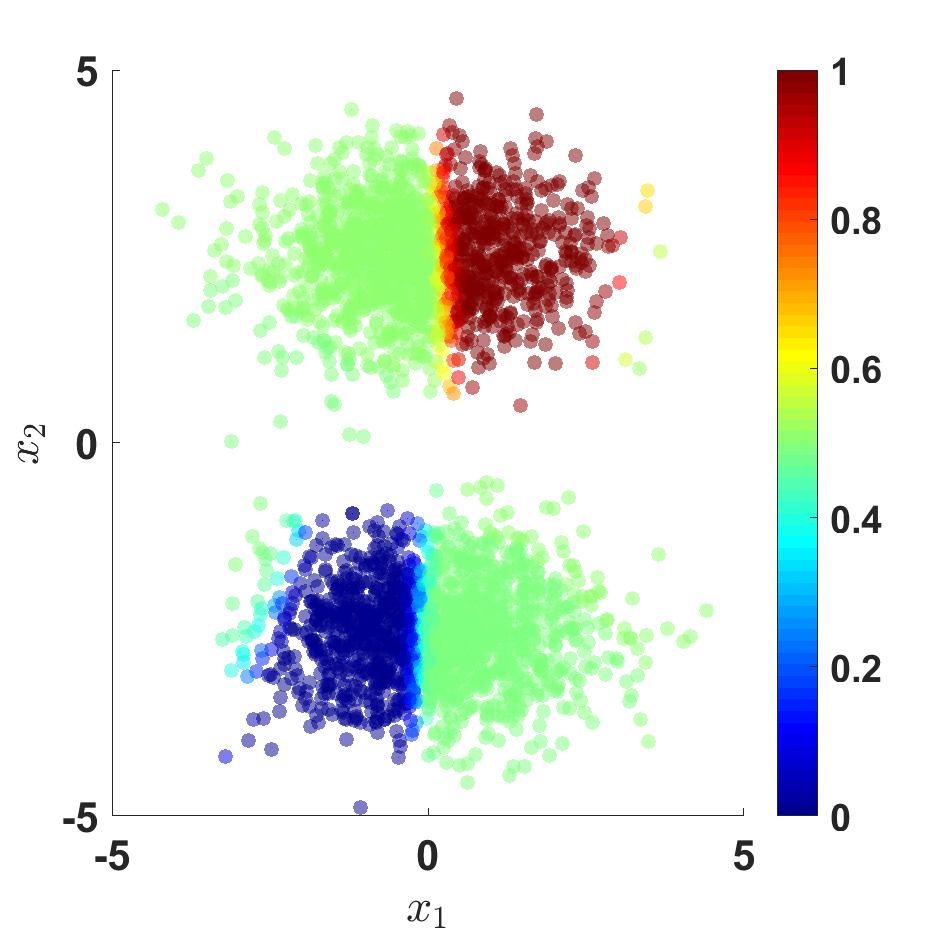}}
    \caption{Example based on \cite[example~3.7]{zhu2009introduction} in two-dimensional feature space $\gX \subset \mathbb{R}^2$ and labels missing at random (MAR). (a) Data with vertically separable classes, where the marginal distribution $p( \vx )$ concentrates as two clusters in the feature space. Note that the commonly assumed `cluster/label' correspondence is not valid here which renders clustering-based methods inappropriate (e.g., label propagation \cite{zhu2002learning}). (b) The MCAR-based semi-supervised learning methods, such as self-training \cite[algorithm~2.4]{zhu2009introduction}, aggressively utilize the unlabeled data, resulting in severely overconfident models and poor classification performance in feature regions with only unlabeled data. (c) The method proposed in Sec.~\ref{sec:method} is capable of learning reliable models for classification.}
\end{figure*}

\begin{itemize}
	\item The label $y$ is missing completely at random (MCAR): The unlabeled and labeled data-generating processes match exactly, i.e.,
	\begin{equation}
	\begin{split}
		p(\ \vx, y \ |\ \lab = 0 \ ) &\equiv p(\ \vx, y  \ |\ \lab = 1 \ ).
	\end{split}
	\label{eq:mcar}
	\end{equation}
	MCAR is a common assumption but can also be highly restrictive. For instance, in a medical diagnosis application, this means that unlabeled features from an unscreened population must statistically match the population of screened patients. When this assumption fails, a semi-supervised learning method may provide severely misleading decisions. For an illustration of this limitation, consider Figures~\ref{fig:example_a} and \ref{fig:example_b}.
	
	\item The label $y$ is missing at random (MAR): The feature distributions may not match each other, i.e.,
	\begin{equation}
	\begin{split}
		p(\ \vx\ |\ \lab = 0 \ ) &\neq p(\ \vx  \ |\ \lab = 1 \ ), \\  
		p(\ y \ |\ \vx, \lab = 0 \ ) &\equiv p(\ y  \ |\ \vx, \lab = 1 \ ),
	\end{split}
	\label{eq:mar}
	\end{equation}
	but the regions of the feature space in which it is possible to discriminate between classes remain invariant (\ref{eq:mar}).
	In the medical diagnosis example, this means that data from the unscreened population can be used in conjunction with screened patient data even though their features differ statistically.
	
	\item The label $y$ is missing not at random (MNAR): Neither features nor conditional class distributions may match each other, i.e.,
	\begin{align*}
		p(\ \vx \ |\  \lab = 0 \ ) &\neq p(\ \vx  \ |\  \lab = 1 \ ), \\  
		p(\ y \ |\ \vx,  \lab = 0 \ ) &\neq p(\ y  \ |\ \vx,  \lab = 1 \ ).
	\end{align*}
	In this case, there is no necessary relation between the labeled and unlabeled data. This effectively invalidates the use of semi-supervised methods as the unlabeled data cannot be utilized to improve a classification.
\end{itemize}
The utilization of unlabeled data, therefore, depends critically on assumptions made about the data-generating distributions underlying labeled and unlabeled data, see also \cite{chawla2005learning}.

\subsection{Semi-supervised learning with reliable uncertainty measure}
For safety-critical applications (e.g., medical diagnosis), the user is interested in not only optimal decisions but also the associated risk analysis. Specifically, the error probability $p( y \neq \widehat{y}\ |\ \vx_*)$ of predicting the label $\widehat{y}$, given a test sample with the feature $\vx_*$, is a central measure of uncertainty. This quantity is unknown and needs to be estimated. If the estimated error probability of a learned classifier matches the true error probability, we say this classifier is reliable or well-calibrated \cite{cohen2004properties}. 
The reliability of the uncertainty measure provided by a classifier is usually evaluated by using a reliability diagram \cite{murphy1977reliability} or calibration score \cite{blattenberger1985separating}. Here we focus on $p( y \neq \widehat{y}\ |\ \vx_*)$ as a measure and consider the following definition:
\begin{definition}
The \emph{reliability} of a classifier is the closeness of its nominal error probability to the error probability obtained out-of-sample test data.    
\end{definition}
It is a non-trivial task to obtain the reliable classifier, even under the supervised settings \cite{niculescu2005predicting,guo2017calibration}. Many models require running post-processing calibration algorithms using an additional, hold-out calibration dataset $\datasetl'$ to produce reliable error probabilities: That is, given a continuous output $f(\vx)$ from a learned classifier, a post-processing method tries to fit a mapping from  $f(\vx)$ to the empirical probability $\widehat{p}(y\ |\ \vx)$ using $\datasetl'$. Such methods include Platt scaling \cite{platt1999probabilistic}, Isotonic regression \cite{zadrozny2001obtaining,zadrozny2002transforming}, and Bayes binning \cite{naeini2015obtaining}. To sum up, calibration is a post-processing technique, which transforms the classification score to a reliable error probability estimate.


To the best of our knowledge, there is very little previous work when it comes to reliable semi-supervised learning. A fundamental problem under MAR is that there is no calibration dataset available for which we can obtain the empirical probability $\widehat{p}(y\ |\ \vx)$ for the unlabeled data $\vx$ that differ from the labeled ones. In summary, for semi-supervised learning under MAR, the utilization of unlabeled data affects not only the classification accuracy, but it also generate impact on the reliability of the learned classifiers.

\section{Related Work}
Relatively few works in the literature address the fundamental assumptions made about the labeling process. Instead, most previous works build upon the 
MCAR (\ref{eq:mcar}) assumption and that information about $y$ for unlabeled features is provided by their proximity to labeled features, e.g. \cite{zhu2009introduction,kingma2014semi,gordon2017bayes}. Transductive methods, such as T-SVM \cite{joachims1999transductive}, S$^3$VM \cite{bennett1999semi}, manifold regularization \cite{belkin2006manifold}, and their successors \cite{wang2012new, li2015towards}, are developed based on the additional assumption that clusters in the feature space correspond to distinct classes. Under this assumption these methods perform well. More recent deep learning-based methods, such as temporal ensembling \cite{laine2016temporal}, mean teacher \cite{tarvainen2017mean}, and adversarial training \cite{springenberg2015unsupervised,chen2016infogan,odena2016semi,miyato2018virtual}, are able to achieve very good testing performance on popular benchmarking datasets. Nevertheless, both classic transductive methods and recent deep learning methods are only applicable to the MCAR scenario and their performance suffer when the training data contains large samples from feature regions that are rarely labeled \cite{oliver2018realistic}. 

As Figure~\ref{fig:example_b} illustrates, the MCAR assumption can lead to severely degraded classifiers with statistically inaccurate, or unreliable, measures of label uncertainty. At best the unreliable measures can be calibrated after learning, using an additional training dataset. From a user perspective, methods that provide reliable uncertainty measures are often more valuable than methods with better predictive performance but poor reliability \cite{cohen2004properties}. In \cite{guo2017calibration}, the authors showed that modern neural networks are poorly calibrated (i.e., the estimated confidence does not match the empirical accuracy rate). To produce `good' probabilistic decisions, most existing approaches are designed for supervised learning and require post-processing of results generated by non-calibrated models \cite{niculescu2005predicting}. In \cite{cohen2004properties}, the authors proved that the Bayes error of a classifier is upper bounded by the error probability of a well-calibrated classifier. Furthermore, the authors showed that calibration of a classifier will not decrease its accuracy. Under the supervised setup, Niculescu-Mizil et.al. \cite{niculescu2005predicting} examined the ability of providing reliable posterior probabilities of ten different classifiers, and tested the effectiveness of two calibration methods, Platt scaling and isotonic regression, on those classifiers. In \cite{moreno2012unifying}, the authors refer the three assumptions of missing data (MCAR, MAR, and MNAR) as the sample selection bias: the training data might be selected in a non-uniform manner from the population. As the authors stated, the techniques for correcting the sample selection bias might fail when a sub-population with specific feature values is not sampled in the training data, which is known as the censorship problem in statistics. More recently, Guo et.al. \cite{he2016deep} discovered that the modern neural networks, such as the ResNet, are poorly calibrated. However, the reason as to why the modern neural networks are miscalibrated remains an open problem.  As stated in \cite{moreno2012unifying}, it is a non-trivial tasks to produce calibrated probabilities for semi-supervised learning algorithms under the MAR assumption.

In this paper, we depart from the conventional but restrictive MCAR assumption and consider semi-supervised learning under MAR. Specifically,
\begin{itemize}
	\item we develop a method that is robust to mismatches between the labeled and unlabeled feature distributions, 
	\item the proposed semi-supervised method is directly applicable using any generative model, with associated supervised learning algorithm, see e.g. \cite{2016-Hastie-ElementsLearning,bishop2016pattern,murphy2012machine},
	\item the proposed method is capable of providing reliable uncertainty measures without post-processing. 
\end{itemize}

\emph{Notation:} The sample mean is $\E_n[\vx] = n^{-1} \sum^n_{i=1}\vx_i$.

\section{Problem Formulation}
We seek a classifier that, given a test point $\vx_*$, is capable of providing predictions of the class label $y_*$ with reliable uncertainty measures under MAR (\ref{eq:mar}). 

\subsection{Optimal Classifier under MAR}
We begin by considering an optimal classifier $\widehat{y}(\vx) \in \gY$ given the unknown data distributions. Without loss of generality, we consider here the standard zero-one loss function, 
\begin{equation}
L\big(\widehat{y}(\vx_*), y_*\big) = \begin{cases} 0 & \text{if $\widehat{y}(\vx_*) = y_*$}, \\
1 & \text{if $\widehat{y}(\vx_*) \neq y_*$},
\end{cases}
\end{equation}
at test sample $\vx_*$ \cite{2016-Hastie-ElementsLearning}.

Under (\ref{eq:mar}), it is impossible to know whether the feature $\vx_* \sim p(\vx\ |\  \lab)$ is drawn under conditions corresponding to $\lab=0$ or $\lab=1$. By considering the sampling scenario to be random $\lab \sim p(\lab)$, the expected loss function under MAR is
\begin{equation*}
	\begin{split}
		& \E[L(y(\vx_*), y_*)] \\
		= &\sum_{\lab \in \{0,1 \}} \sum_{y \in \gY} p(\lab) \int_\gX  L\big(\widehat{y}(\vx), y\big)p(y\ |\ \vx)p(\vx\ |\ \lab) d\vx \\
	\end{split}
\end{equation*}
where $p(y\ |\ \vx) \equiv p(y\ |\ \vx, \lab)$ for all $\lab$. The optimal classifier is then given by
\begin{equation}
\begin{split}
\widehat{y}(\vx_*) &= \underset{y \in \gY}{\arg\max}\ p(\vx_*\ |\ y)p(y), 
\label{eq:optimal_classifier}
\end{split}
\end{equation}
where the marginalized distributions are
\begin{equation}
\begin{split}
p(\vx\ |\ y) &= \sum_{\lab \in \{0,1\} } p(\lab) \: p(\vx\ |\ y,\lab), \\ 
p(y) &= \sum_{\lab\in \{0,1\}}  p(\lab) \: p(y\ |\ \lab).
\end{split}
\label{eq:marginals}
\end{equation}
The uncertainty of the classifier output is quantified by the conditional error probability, which for notational simplicity we denote as
\begin{equation}
\begin{split}
p_e(\vx_*) \equiv p( y \neq \widehat{y}\ |\ \vx_*) = 1 - \frac{p(\vx_*\ |\ \widehat{y})p(\widehat{y})}{\sum_y p(\vx_*\ |\ y)p(y)},
\label{eq:error_prob}
\end{split}
\end{equation}
which also depends on (\ref{eq:marginals}).

\subsection[Learning models]{Learning a Generative Model}
Let
$$q = \big\{ \: q(\vx\ |\ y, \lab), \: q(y\ |\ \lab) \: \big\} \; \in \; \mathcal{Q},$$
denote a model of $p(\vx\ |\ y,\lab)$ and $p(y\ |\ \lab)$ that belongs to some model class $\mathcal{Q}$. Consider learning $q$ using the maximum likelihood approach, which we can express as
\begin{equation}
\begin{split}
\max_{q \in \mathcal{Q}}\quad &\E_{n_1}\Big[ \ln q(\vx\ |\ y,\lab=1)q(y\ |\ \lab=1)\Big] \; \\
&+ \; \frac{n_0}{n_1} \: \E_{n_0}\left[\ln \sum_{y\in \mathcal{Y}}q(\vx\ |\ y,\lab=0)q(y\ |\ \lab=0)\right],
\end{split}
\label{eq:ML}
\end{equation}
where $n_0$ and $n_1$ denote the number of unlabeled and labeled samples used to form the averages. Then the learned $q$ is used to form an approximation of (\ref{eq:marginals})  for use in (\ref{eq:optimal_classifier}) and (\ref{eq:error_prob}). Without imposing further constraints on $q$, the second term in (\ref{eq:ML}) has no unique maximizer and semi-supervised learning of (\ref{eq:marginals}) is impossible. 

MCAR (\ref{eq:mcar}) imposes the following constraints  $q(\vx\ |\ y, \lab= 1) \equiv q(\vx\ |\ y, \lab = 0)$ and $q(y\ |\ \lab=1) \equiv q(y\ |\ \lab=0)$. Then (\ref{eq:ML}) can be solved  approximately using the Expectation Maximization approach, which finds a local maximum, see \cite[algorithm~3.3]{zhu2009introduction}. In this approach $n_0 = |\dataset_0|$ and $n_1 = |\dataset_1|$, and the missing labels are represented by introducing $n_0$ latent variables which can be very large. An alternative, more tractable approach is label-sampling, which imposes the constraint indirectly and obviates the need for introducing latent variables.  In this approach, all unlabeled features are first assigned labels and then augmented with the labeled dataset so that  $n_1 = |\dataset_1| + |\dataset_0|$ and $n_0 = 0$ in (\ref{eq:ML}). This is achieved by learning an initial model of $p(y\ |\ \vx)$ using $\dataset_1$ alone and then predicting $y$ for each $\vx \in \dataset_0$, see \cite[algorithm~2.4]{zhu2009introduction}.

Under MAR, however, the above constraints are invalid and imposing them may adversely affect the learned models as illustrated in Figure~\ref{fig:example_b}. Consequently, the corresponding learned error  probability $q_e(\vx)$ will be inaccurate, resulting in an overconfident and unreliable classifier. That is, the gap $p_e(\vx) - q_e(\vx)$
will be large in regions of $\mathcal{X}$ where there is little labeled data, despite reporting a low error probability $q_e(\vx)$. 

Our task is to formulate a tractable label-sampling approach to tackle (\ref{eq:ML}) that imposes less restrictive constraints and is capable of providing reliable probabilities $q_e(\vx)$ under MAR.

\section{Learning Approach under MAR}
\label{sec:method}
Under (\ref{eq:mar}), datasets $\datasetu$ and $\datasetl$ may cover different regions of the feature space $\gX$ according to the distributions $p(\vx\ |\ \lab=0)$ and $p(\vx\ |\ \lab=1)$, respectively. Certain feature regions that are sampled in $\datasetl$ intersect with those sampled in $\datasetu$. It is in these feature regions that unlabeled features are known to be associated with certain labels. We proceed to formalize this principle into a label-sampling approach to (\ref{eq:ML}).

\subsection{Regions of Label-Informative Features}

Suppose we can learn initial generative models 
\begin{equation}
\begin{cases}
\modelinit(\vx\ |\ y, \lab=1) \text{ and } \modelinit(y\ |\ \lab=1) \text{ from } \dataset_1, \\
\modelinit(\vx\ |\ \lab=0) \text{ from } \dataset_0,
\end{cases}
\label{eq:initialmodel}
\end{equation}
as discussed below. We expect a high accuracy of $\modelinit(\vx\ |\ \lab=0)$ since $|\datasetu|$ is very large. 

Now consider the log-likelihood ratio $\ln [\modelinit(\vx\ |\ y, \lab=1)/\modelinit(\vx\ |\ \lab=0)]$ for a given feature $\vx$. When this statistic is zero, the model cannot discriminate whether $\vx$ is drawn from a labeled or an unlabeled distribution. This fact enables a partition of the feature space $\gXinfo \subseteq \gX$, where
\begin{equation}
	\gXinfo = 
	\left\{ \: \vx \in \gX \: : \:
	\ln \frac{\modelinit(\vx\ |\ y , \lab = 1)}{\modelinit(\vx\ |\ \lab = 0)} > \kappa, \: \exists y \in \gY \right\}.
\label{eq:region}
\end{equation}
All features in $\gXinfo$ are statistically more similar to labeled than unlabeled data, and are therefore label-informative. Testing whether a feature belongs to $\gXinfo$ corresponds to a likelihood ratio test with a threshold $\kappa$ (set to zero by default). 

\emph{Remark:} In safety-critical applications, where cost of miss-classification is high, the region $\gXinfo$ can be made smaller and more reliable by increasing $\kappa \geq 0$.

\subsection{Using Deep Generative Models Under MAR}
There are many ways to learn the initial models in (\ref{eq:initialmodel}), including density estimation and mixture models. However, these classical methods typically do not work well with high-dimensional data, such as images. The recent advance of deep learning provide a new tool, namely the deep generative model, for learning initial models (\ref{eq:initialmodel}) \cite{kingma2014semi, maaloe2016auxiliary, makhzani2015adversarial}. In this section, we give an example of using a deep generative model for semi-supervised learning under MAR. Specifically, we consider the cases that when the numbers of labeled data are unbalanced for different classes.

The generative models are learned here in two steps: dimensionality reduction followed by density estimation. In the dimensionality reduction step, a feature extraction method is used to find a low-dimensional latent representation $\widetilde{\vx}$ of the data feature $\vx$. The representation can be formed using a nonlinear deterministic mapping,
\begin{equation}
    \widetilde{\vx} = f_{\vphi}(\vx),  \label{eq:latent}
\end{equation}
where $f_{\vphi}(\cdot)$ is an encoder network with parameters $\vphi$. Alternatively, one can use a probabilistic mapping,
\begin{equation}
    \widetilde{\vx} \sim p(\widetilde{\vx}\ |\ \vx),  \label{eq:latent_prob}
\end{equation}
where the posterior distribution $p(\widetilde{\vx}\ |\ \vx)$ is approximated via the encoder network of a deep generative model, $p(\widetilde{\vx}\ |\ \vx) \approx q_\vphi(\widetilde{\vx} |\ \vx)$ parameterized by $\vphi$. See \cite{bengio2013representation} and \cite{kingma2013auto} for a review of methods and further details beyond the scope of this paper.

In the density estimation step, we learn generative models for the high-dimensional feature space $\mathcal{X}$ via the low-dimensional space $\widetilde{\mathcal{X}}$. That is, using the compressed representation (\ref{eq:latent}), we learn a low-dimensional density model of the distribution of $\vx$ for each labeled class as well as the unlabeled features.

\subsection{Selective Label-Sampling Approach}

Since all unlabeled features $\vx \in \datasetu$ that belong to $\gXinfo$ are label informative, we apply the label-sampling approach to this subset of features. That is,
\begin{equation*}
	\begin{split}
		\text{if } \vx \in \datasetu \cap \gXinfo  \: : \: \text{assign class } y \sim \modelinit(y\ |\ \vx, \lab=0),
	\end{split}
\end{equation*}
where we use (\ref{eq:mar}) to obtain
\begin{equation*}
	\begin{split}
		\modelinit(y\ |\ \vx, \lab=0) &\equiv  \modelinit(y\ |\ \vx, \lab=1)\\
		&\propto \modelinit(\vx\ |\ y,\lab=1)\modelinit(y\ |\ \lab=1).
	\end{split}    
\end{equation*}
Thus label uncertainty learned from $\datasetl$ is preserved to the sampled data. The resulting pairs $(\vx, y)$ are augmented with $\datasetl$ to form a set $\dataset'$ so that $n_1 = |\dataset'|$ in (\ref{eq:ML}).

By contrast, the remaining unlabeled features, $\vx \in \datasetu \cap \gXinfo^c$, are not informative of $y$ and form a set $\dataset''$ so that $n_0 = |\dataset''|$ in (\ref{eq:ML}). Based on these features, a robust model of $p(\vx\ |\ y, \lab = 0)$ should be class independent. Moreover, the principle of insufficient reason dictates a uniform distribution for the prior label probability. That is, we impose the constraints
\begin{equation}
\begin{split}
q(\vx\ |\ y, \lab = 0)  \; &\equiv \;  q(\vx\ |\ \lab = 0) \\
q(y\ |\ \lab=0) \; &\equiv \;  \frac{1}{|\gY|}
\end{split}
\label{eq:constraint}
\end{equation}
in (\ref{eq:ML}).

In summary, we solve (\ref{eq:ML}) using $\dataset'$ and $\dataset''$ along with the constraints (\ref{eq:constraint}).  Then, we use the estimate $q(\lab) = w \ell + (1-w)(1-\ell)$, where $w = |\dataset'|/(|\dataset'| + |\dataset''|)$, to form an approximation of (\ref{eq:marginals}). For a test sample $\vx_*$, the resulting classifier is given by
$$
\widehat{y}(\vx_*) = \underset{y \in \gY}{\arg\max}\; q(\vx_*\ |\ y)q(y)
$$
and the learned error probability is
\begin{equation}
q_e(\vx_*) = 1 - \frac{q(\vx_*\ |\ \widehat{y})q(\widehat{y})}{\sum_y q(\vx_*\ |\ y)q(y)}.
\label{eq:errormodel}
\end{equation}
The proposed approach prevents misfitting models to features $\dataset''$ that lack sufficient label information under MAR, which would otherwise lead to overconfident classifiers and unreliable error probabilities $q_e(\vx)$. The method is summarized in Algorithm~\ref{alg:code} and can be implemented using \emph{any} generative model and learning algorithm of choice. As illustrated in Figure~\ref{fig:example_c}, the proposed methods avoids overconfident predictions as compared to Figure~\ref{fig:example_b}, while maintaining accurate predictions in regions that contain labeled data.
 
\begin{algorithm}[tb]
	\caption{Proposed Semi-Supervised Method}
	\label{alg:code}
	\begin{algorithmic}
		\STATE {\bfseries Input:} $\datasetl$ and $\datasetu$
		\STATE Learn $\modelinit(\vx\ |\ y,\lab=1)$ and $\modelinit(y\ |\ \lab=1)$ using $\datasetl$\;
		\STATE Learn $\modelinit(\vx\ |\ \lab=0)$ using $\datasetu$\;
		\STATE Set $\dataset':=\datasetl$ and $\dataset''=\datasetu \cap \gXinfo^c$\;
		\FOR{$\vx \in \datasetu \cap \gXinfo$}
		\STATE 	Draw $y \sim \modelinit(y\ |\ \vx,\lab=0)$\;
		\STATE $\dataset':= \dataset' \cup \{(\vx, y)\}$\;
		\ENDFOR
		\STATE Learn model $q$ via (\ref{eq:ML}) with $n_1 = |\dataset'|$ and $n_0 = |\dataset''|$, imposing (\ref{eq:constraint})\;
		\STATE Let $w = |\dataset'|/(|\dataset'| + |\dataset''|)$\;
		\STATE {\bfseries Output:} $q(\vx\ |\ y) = w q(\vx\ |\ y,\lab=1) + (1-w)q(\vx\ |\ \lab=0)$ and $q(y)=wq(y\ |\ \ell=1)+(1-w)\frac{1}{|\gY|}$
	\end{algorithmic}
\end{algorithm}

\section{Experimental Results}
To illustrate the proposed approach with real data, we consider datasets of handwritten digit and cloth classification where the labels are missing at random. Our focus here is to compare the reliability of the learned error probabilities $q_e(\vx)$ with the analogous MCAR-based semi-supervised and supervised methods. That is, we assess the extent to which the gap $p_e(\vx) - q_e(\vx)$ is statistically small and non-negative for a large test dataset, using reliability diagrams \cite{murphy1977reliability,degroot1983comparison}.

To provide a reliable classifier, the model class $\mathcal{Q}$ in (\ref{eq:ML}) must be suitably adapted to the given feature type. One option is multivariate kernel density models \cite{silverman2018density}, where each sample is represented by a choice of kernel functions and bandwidth matrices. Together the kernel functions give a density estimate for an arbitrary testing sample. While this method is able to provide highly adaptive data models, it requires data sizes that increase exponentially with the feature dimension. A second option is to use finite mixture models \cite{bishop2016pattern}, for example Gaussian mixture models (GMM). The parameters in GMM (weights, means, and covariance matrices) can be learned using the expectation maximization algorithm or a variational Bayesian approach which enables regularized learning with many mixture components. A third option is infinite mixture models, such as the Dirichlet process mixture models (DPM) \cite{gorur2010dirichlet} with Markov chain Monte Carlo (MCMC) techniques. These models provide adaptive data models and can tackle increasing feature dimensions better than kernel density methods. Their implementation, however, also requires a lot more work.

For clarity and ease of implementation in the examples below, we consider learning Gaussian mixture models using a variational Bayes method (VB-GMM) in conjunction with the variational auto-encoder (VAE) \cite{kingma2013auto}. The model class $\mathcal{Q}$ is applied to the proposed method as well as the corresponding MCAR-based semi-supervised method. For reference, we also consider a generative supervised method using only $\datasetl$. For the initial models $\modelinit(\vx\ |\ \lab=0)$ and $\modelinit(\vx\ |\ y, \lab=1)$ in Algorithm~\ref{alg:code}, we also use VB-GMM.

\subsection{MNIST dataset}

\begin{figure*}
	\centering
	\subcaptionbox{Labeled (colored) and unlabeled (black) training data\label{fig:mnist_data}}[.45\linewidth]
	{\includegraphics[width=0.3\textwidth]{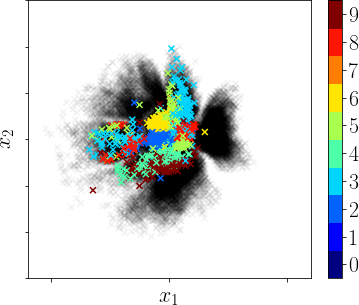}}%
	\hfill
	\subcaptionbox{Log-likelihood ratio $\ln \modelinit(\vx|y, \lab=1)/ \modelinit(\vx| \lab=0)$ using the maximizing label $y$.  \label{fig:mnist_clrt}}[.45\linewidth]
	{\includegraphics[width=0.3\textwidth]{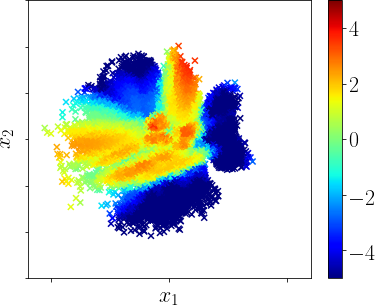}}%
	\caption{(a) MNIST data and (b) log-likelihood ratio with maximizing label across feature space $\gX$. The region in which the ratio falls below $\kappa=0$ equals $\gXinfo$ in (\ref{eq:region}).}
\end{figure*}

\begin{figure*}
	\centering
	\subcaptionbox{MAR semi-supervised learning\label{fig:mnist_pe_1}}[.32\linewidth]
	{\includegraphics[width=0.3\textwidth]{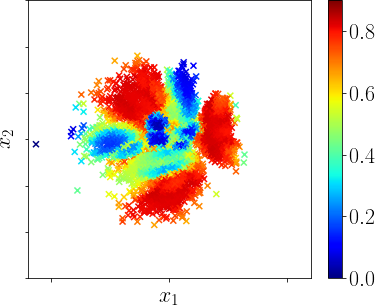}}%
	\hfill
	\subcaptionbox{MCAR semi-supervised learning \label{fig:mnist_pe_2}}[.32\linewidth]
	{\includegraphics[width=0.3\textwidth]{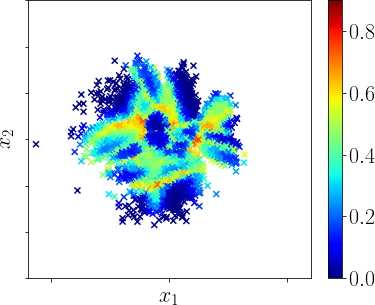}}%
	\hfill
	\subcaptionbox{Supervised learning\label{fig:mnist_pe_3}}[.32\linewidth]
	{\includegraphics[width=0.3\textwidth]{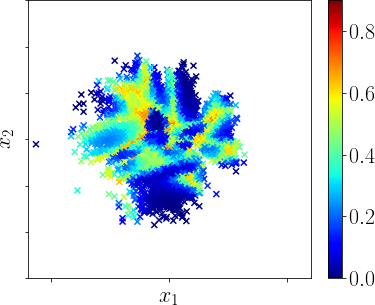}}%
	\caption{Error probability output $q_e(\vx_*)$ across testing data. The proposed method yields a reliably high $q_e(\vx_*)$ for testing samples with features outside of $\gXinfo$  (red region where there is no labeled data, see Fig.~\ref{fig:mnist_data}). \label{fig:mnist_pe}}
\end{figure*}

\begin{figure*}
	\centering
	\subcaptionbox{MAR semi-supervised learning\label{fig:mnist_empirical_1}}[.32\linewidth]
	{\includegraphics[width=0.3\textwidth]{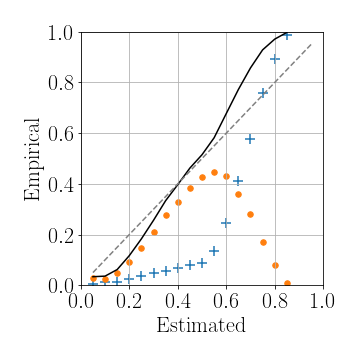}}%
	\hfill
	\subcaptionbox{MCAR semi-supervised learning\label{fig:mnist_empirical_2}}[.32\linewidth]
	{\includegraphics[width=0.3\textwidth]{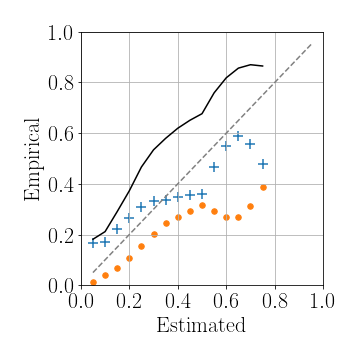}}%
	\hfill
	\subcaptionbox{Supervised learning\label{fig:mnist_empirical_3}}[.32\linewidth]
	{\includegraphics[width=0.3\textwidth]{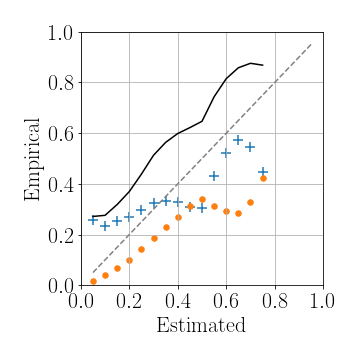}}%
	\caption{Reliability diagrams for MNIST data: Error probability outputs $q_e(\vx_\star)$ versus empirical error probabilities $\widehat{p}_e(\vx_\star)$ (solid) for testing data. A curve that exceeds the ideal 45 degree diagonal line corresponds to an overconfident classifier. The empirical probabilities are decomposed into two contributions: from rarely labeled features (crosses), that are highly associated with labels `0', `1', and `7', and remaining features (dots), respectively. \label{fig:mnist_empirical}}
\end{figure*}

\begin{figure*}
	\centering
	\subcaptionbox{MAR semi-supervised learning\label{fig:fmnist_empirical_1}}[.32\linewidth]
	{\includegraphics[width=0.3\textwidth]{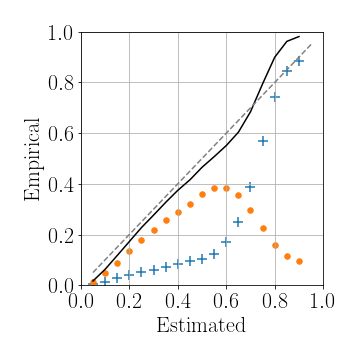}}%
	\hfill
	\subcaptionbox{MCAR semi-supervised learning\label{fig:fmnist_empirical_2}}[.32\linewidth]
	{\includegraphics[width=0.3\textwidth]{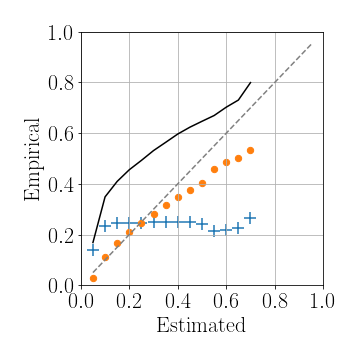}}%
	\hfill
	\subcaptionbox{Supervised learning\label{fig:fmnist_empirical_3}}[.32\linewidth]
	{\includegraphics[width=0.3\textwidth]{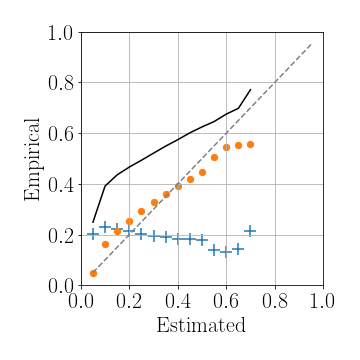}}%
	\caption{Reliability diagrams for Fashion MNIST data: Error probability outputs $q_e(\vx_\star)$ versus empirical error probabilities $\widehat{p}_e(\vx_\star)$ (solid) for testing data.  Empirical probabilities are decomposed into contributions from rarely labeled features (crosses) that are highly associated with labels {`T-shirt' and `Trouser'}, and from remaining features (dots)} \label{fig:fmnist_empirical}
\end{figure*}

In the first experiment, we consider the MNIST dataset which consists of $28\times28$ gray-scale images of handwritten digits, labeled $\gY = \{0,1, \dots, 9\}$ \cite{lecun1998gradient}. For the purpose of visualization, we use two-dimensional features $\vx$ for each image, obtained using a deep variational autoencoder (VAE) \cite{kingma2014semi}. We consider the MAR scenario (\ref{eq:mar}) in which the labeled data in $\datasetl$ provide few samples in the feature regions that are highly predictive of labels $0,1$ and $7$. We use $|\datasetl| = 1~000$ and $|\datasetu| = 59~000$ samples. See Figure \ref{fig:mnist_data}, which clearly illustrates two sources of label uncertainty: regions that lack labeled data and regions in which classes overlap.

To illustrate the first source, we plot the log-likelihood ratio (\ref{eq:region}) in Figure~\ref{fig:mnist_clrt}. This forms the basis for $\gXinfo$ and the error probability output $q_e(\vx)$ is shown for test data in Figure~\ref{fig:mnist_pe_1}, which qualitatively reflects the two sources of label uncertainty. 
By contrast, the output of the MCAR-based label-sampling approach in Figure~\ref{fig:mnist_pe_2} only reflects uncertainty in regions where classes overlap or closely neighbour each other. The MCAR-based label-sampling approach drastically fails to represent the uncertainty due to the shift of labeled and unlabeled data distribution, i.e., $p(\vx\ |\ \lab = 0 ) \neq p( \vx\ |\ \lab = 1 )$. 
For reference, Figure~\ref{fig:mnist_pe_3} shows the output of the supervised method, which uses only $\datasetl$. We see that the MCAR-based method does not perform significantly different. A comparison of these figures illustrates the ability of the proposed semi-supervised method to yield reliable predictions under MAR.

We now compare the reliability of the learned models for test data, using the reliability diagram \cite{murphy1977reliability,degroot1983comparison,guo2017calibration}. That is, we compare the error probability output $q_e(\vx_\star)$ of the models with their empirical error probabilities $\widehat{p}_e(\vx_\star)$ using actual test data with 10~000 samples. If $q_e(\vx_\star)$ systematically underestimates $\widehat{p}_e(\vx_\star)$, the classifier is based on an unreliable model. Figure~\ref{fig:mnist_empirical} corroborates the previous illustration and shows how the proposed method reliably follows the empirical error probability fairly, whereas the analogous MCAR alternatives systematically underestimate it and thus provides unreliable label predictions.

\subsection{Fashion MNIST Dataset}
In the second experiment, we use the more challenging Fashion-MNIST dataset \cite{xiao2017fashion}, which contains $28 \times 28$ gray-scale images with $|\gY| = 10$ classes of fashion items (ranging from T-shirts to ankle boots). We construct features $\vx$ for each image using the deep VAE technique above \cite{kingma2014semi}. The constructed features are chosen to be five-dimensional here, which provides sufficient separation between the classes for meaningful classification. We consider a MAR scenario (\ref{eq:mar}) in which the labeled features in $\datasetl$ provide few samples in the that are highly associated with the labels `T-shirt' and `Trouser'. We use $|\datasetl| = 1~000$ and $|\datasetu| = 59~000$ samples.  As in the previous experiment, we apply our proposed method in Algorithm~\ref{alg:code} using VB-GMM. 

Again the reliability of the proposed semi-supervised learning method is corroborated by comparing the error probability output $q_e(\vx_\star)$ with the empirical error probability $\widehat{p}_e(\vx_\star)$ using the Fashion-MNIST test data with 10~000 samples. In Fig.~\ref{fig:fmnist_empirical_1}, we see that the error probability output approximates the empirical error fairly well, following a near diagonal line. By contrast, Figs.~\ref{fig:fmnist_empirical_2} and \ref{fig:fmnist_empirical_3} show that the analogous MCAR-based method and the supervised method systematically underestimate the empirical error, resulting in unreliable classifications.

\section{Conclusion}
We have developed a semi-supervised learning method that is capable of providing reliable label predictions in cases where the labeled training data is missing at random. This is a \emph{less restrictive} scenario than the missing completely at random assumption upon which many existing semi-supervised method are built. Such methods produce unreliable predictions of the label when the assumption is violated. One advantage of the proposed method is its wide applicability using any generative model with an associated supervised learning algorithm.  We illustrated the capability of the approach to provide reliable uncertainty measures when labels are missing at random, using both handwritten digit and cloth classification. Further research may evaluate the method together more advanced models and training algorithms.

\bibliographystyle{IEEEtran}
\bibliography{reference}

\end{document}